\documentclass{bmvc2k}
\usepackage{amssymb}
\usepackage{amsmath}
\usepackage{float}
\usepackage{pifont}
\usepackage{booktabs}
\usepackage{nicematrix}
\newcommand{\cmark}{\ding{51}}%
\newcommand{\xmark}{\ding{55}}%

\title{Open-world Text-specified Object Counting}

\addauthor{Niki Amini-Naieni}{niki.amini-naieni@eng.ox.ac.uk}{1}
\addauthor{Kiana Amini-Naieni}{kamininaieni@ucdavis.edu}{2}
\addauthor{Tengda Han}{htd@robots.ox.ac.uk}{1}
\addauthor{Andrew Zisserman}{az@robots.ox.ac.uk}{1}

\addinstitution{
 Visual Geometry Group (VGG),\\
 University of Oxford, UK
}
\addinstitution{
 University of California, Davis, USA
}

\runninghead{Amini-Naieni et al.}{Open-world Text-specified Object Counting}

\def\etal{\emph{et al}\bmvaOneDot}

\begin{document}

\maketitle

\begin{abstract}
Our objective is open-world object counting in images, where the target object class is specified by a text description. To this end, we propose {\em CounTX}, a class-agnostic, single-stage model using a transformer decoder counting head on top of pre-trained joint text-image representations. CounTX is able to count the number of instances of any class given only an image and a text description of the target object class, and can be trained end-to-end. 
In addition to this model, we make the following contributions: (i) we compare the performance of CounTX to prior work on open-world object counting, and show that our approach exceeds the state of the art on all measures on the FSC-147~\cite{m_Ranjan-etal-CVPR21} benchmark for methods that use text to specify the task; (ii) we present and release FSC-147-D, an enhanced version of FSC-147 with text descriptions, so that object classes can be described with more detailed language than their simple class names. FSC-147-D and the code are available at \href{https://www.robots.ox.ac.uk/~vgg/research/countx}{https://www.robots.ox.ac.uk/\textasciitilde vgg/research/countx}.

\end{abstract}

\raggedbottom
\section{Introduction}
\label{sec:intro}
The goal of object counting is to estimate the number of relevant objects in an image. Traditional object counting methods focus on counting objects of a specific class of interest \cite{10.1007/978-3-031-19821-2_11, artetaCountingWild2016, mundhenkLargeContextualDataset2016a, doi:10.1080/21681163.2016.1149104}. These techniques are well-suited for solving particular problems, but they cannot operate in {\em open-world} settings, where the class of interest is not known beforehand and can be arbitrary.

One approach to open-world object counting is class-agnostic few-shot object counting~\cite{luClassAgnosticCounting2019, yangClassagnosticFewshotObject2021}. In this setting, a user specifies the class of interest at inference time with one or more visual exemplars. These exemplars take the form of bounding boxes over different instances of the object in the image. Although class-agnostic few-shot object counters can be deployed in open-world scenarios, they require a human to provide the visual exemplars during inference. An alternative approach is for the object of interest to be specified by text, rather than by visual exemplars. This is the approach proposed recently by Xu \etal~\cite{Xu2023ZeroshotOC}, where the objects to be counted can be specified by an \emph{arbitrary} class name at inference time. This is implemented as a two-stage process where first the text is used to select visual exemplars in the image, and then the counting proceeds using these exemplars as in the standard few-shot setting.

\begin{figure*}
\centering
\includegraphics[width=0.8\textwidth]{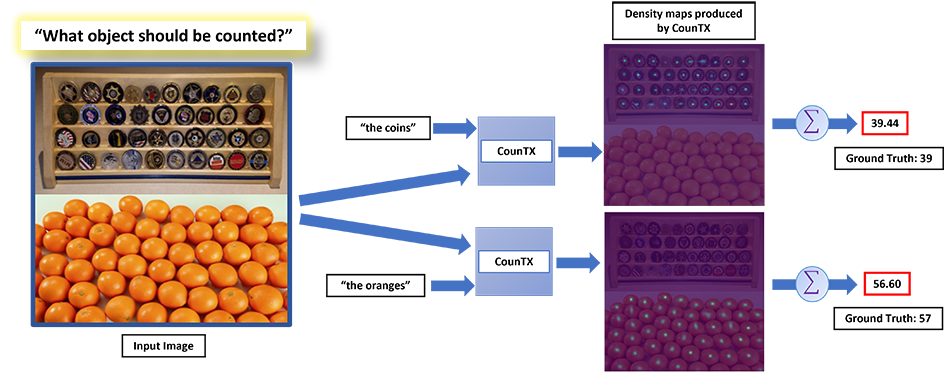}
\vspace{-3mm}
   \caption{
CounTX estimates object counts directly from an image and a response to the question ``what object should be counted?''. In this example, two text inputs are used to predict the object counts of different objects in the same image. Note, no visual exemplars are required at any stage. 
}
\label{fig:short}
\end{figure*}

In this paper, we propose {\em CounTX}, a {\em single-stage} open-world image counting model, where the objects to be counted are specified by a textual description at inference time. CounTX (pronounced ``Count-text") does not use visual exemplars, and can be trained end-to-end. It can count objects of any class, even ones unseen during training. CounTX exceeds the performance of the two-stage approach of~\cite{Xu2023ZeroshotOC} on all measures on FSC-147, a standard counting benchmark. Figure \ref{fig:short} shows an example of its use.

The key idea is to benefit from the availability of image encoders that have been pre-trained for a joint text-image encoding using large scale image-text paired data, such as CLIP~\cite{Radford2021LearningTV}. Taking inspiration from Chang \etal~\cite{Liu2022CounTRTG}, that attention mechanisms can be used to model the similarity between image patches and other inputs, we use a transformer decoder to determine the similarity between the text encoding of the target object description and the spatial map of the image. This generates a density map that is then decoded to the image resolution and used for counting. We also investigate whether it is better to freeze or fine-tune the image or text encoder when training the counting head of the model. 


In short, we make three contributions: \emph{First}, we develop CounTX, an open-world counting model that accepts an image and an arbitrary object class description, and directly uses these inputs to predict the object count. We are the first to tackle this open-world counting problem using a single-stage approach, without relying on an exemplar-based counting model; \emph{Second}, we augment the FSC-147~\cite{m_Ranjan-etal-CVPR21} dataset with class descriptions and release the modified dataset, FSC-147-D, for future research; \emph{Third}, we verify the effectiveness of our model and training procedure on the FSC-147 dataset through both quantitative and qualitative results. CounTX significantly improves on both the validation set and the test set performance 
 of~\cite{Xu2023ZeroshotOC}, the only prior work on open-world text-specified object counting.
\section{Related Work}
\paragraph{Class-specific Object Counting.}
Class-specific object counting focuses on counting objects of a specific category \cite{10.1007/978-3-031-19821-2_11, artetaCountingWild2016, mundhenkLargeContextualDataset2016a, doi:10.1080/21681163.2016.1149104}. There are two main approaches to this task: detection-based methods \cite{Barinova2010OnDO, 10.1007/s11263-011-0439-x, Hsieh2017DroneBasedOC} and regression-based methods \cite{Arteta2014InteractiveOC, artetaCountingWild2016, Cho1999ANC, Kong2006AVI, Lempitsky2010LearningTC, Marana1997EstimationOC, doi:10.1080/21681163.2016.1149104}. While detection-based methods rely on an object detector to produce bounding boxes that can be enumerated to predict the object count, regression-based methods directly map the input image to a continuous scalar estimate of the object count.



\paragraph{Class-agnostic Object Counting.}  The goal of class-agnostic object counting is to count the instances of an arbitrary class in an image given a number of visual exemplars at the time of inference \cite{Arteta2014InteractiveOC, Gong2022ClassAgnosticOC, Liu2022CounTRTG, luClassAgnosticCounting2019, 10.1007/978-3-031-20044-1_20, m_Ranjan-etal-CVPR21, Shi2022RepresentCA, yangClassagnosticFewshotObject2021, You_2023_WACV, low_shot, Lin_2022_BMVC}. Class-agnostic object counters require users to provide a positive number of exemplars to specify the class of the object to count. For instance, CounTR uses a two-stream transformer-based architecture to model the similarity between image patches and visual exemplars \cite{Liu2022CounTRTG}. While CounTR accepts any number of visual exemplars (zero or more), if the user provides zero exemplars, CounTR will default to counting instances of the {\em dominant} class. Although CounTX also uses a two-stream transformer-based architecture, it does not require visual exemplars for class specification. 
There are also general object counting models that do not use exemplars \cite{Hobley2022LearningTC, 10.1007/978-3-031-26316-3_5}. However, they do not allow the user to specify the class of interest. Instead, like CounTR with zero exemplars, these algorithms count instances of the dominant class in the image. 

\paragraph{Text-specified Object Counting.}  The aim of text-specified object counting is to count objects of an arbitrary class in an image given only the class description. Xu \etal~\cite{Xu2023ZeroshotOC} are the first to propose a method for this task. Unlike CounTX, the technique presented by Xu \etal does not produce object counts directly. Instead, it proposes optimal visual exemplars for use by existing class-agnostic object counting networks such as FamNet \cite{m_Ranjan-etal-CVPR21}, BMNet \cite{Shi2022RepresentCA}, BMNet+ \cite{Shi2022RepresentCA}, and Xu \etal's own architecture, all already trained with annotated visual exemplars. Xu \etal  refer to their method as `zero-shot' as it avoids the user inputting the visual exemplars. In contrast, CounTX eliminates the need for annotated visual exemplars altogether, and also accepts a more detailed specification of the target object to count (rather than simply using a class name).

\section{Method}
We consider the problem of open-world object counting in images, where the target object class is specified by a text description. In this setting, classes unseen during training may be encountered during inference.

\subsection{Overview}
Given a training set $\mathcal{D}_{train} = \{(X_{1}, t_{1}, Y_{1}), \dots, (X_{N}, t_{N}, Y_{N})\}$, each $X_{i} \in \mathbb{R}^{H \times W \times 3}$ is a training image with a tokenized class description $t_{i} \in \mathbb{R}^{n}$ and a binary map $Y_{i} \in \mathbb{R}^{H \times W}$ with a one at the center of each object in $X_{i}$ described by $t_{i}$ and zeros at all other entries. As shown in Eq.~\ref{count}, the entries of $Y_{i}$ can be summed to obtain the count of objects in $X_{i}$ described by $t_{i}$:
\begin{equation}\label{count}
Count(X_{i}, t_{i}) = \sum_{p, q}(Y_{i})_{p, q}
\end{equation}
where $p,q$ specify the pixel index.

Our goal is to develop an open-world object counter and to train it on $\mathcal{D}_{train}$ such that it generalizes well to $\mathcal{D}_{test}$, a held-out test set of images with classes not in $D_{train}$. To achieve this, we introduce CounTX, a novel transformer-based architecture that directly determines a density map from each input image and class description. This density map can be summed to estimate the object count. In section \ref{arch}, the inspiration for the CounTX architecture is explained, and each of its modules are described in detail. In section \ref{train}, the motivation and methods used for pre-training and fine-tuning CounTX are outlined.

\subsection{Architecture}\label{arch}

\begin{figure*}
\centering
\includegraphics[width=1\textwidth]{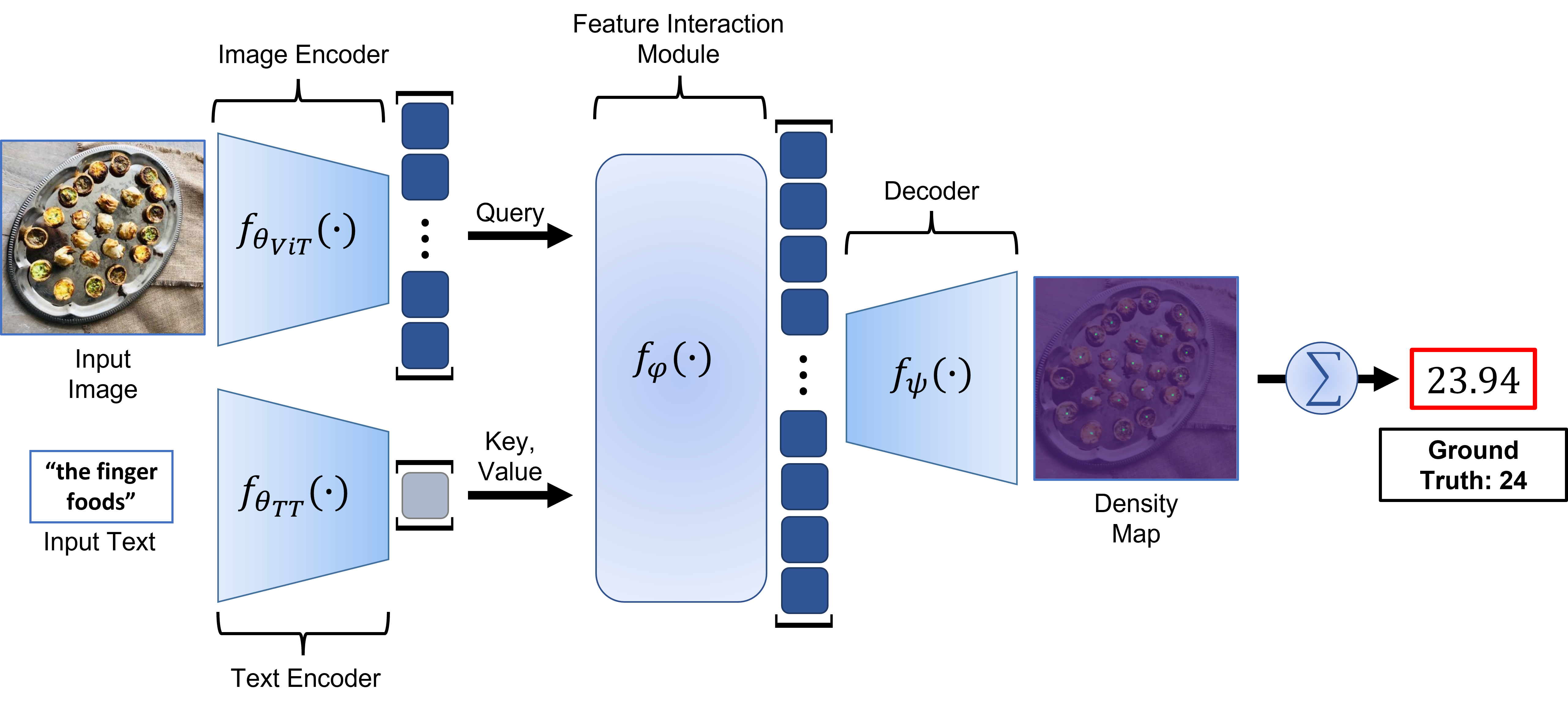}
\vspace{-7mm}
   \caption{The CounTX architecture. The input image and class description are encoded by a vision transformer and a text transformer respectively. The image features are then passed to the feature interaction module to compute the query vectors, and the text feature is passed in to compute the key and value vectors. The output of the feature interaction module is reshaped to a spatial feature map that is upsampled in the decoder module to produce a density map with the same height and width as the input image and entries that sum to the object count.}
\label{fig:arch}
\vspace{-2mm}
\end{figure*}

In this section, we describe the CounTX architecture,  illustrated in Figure~\ref{fig:arch}. CounTX is inspired by the class-agnostic object counting framework, CounTR, of Chang \etal \cite{Liu2022CounTRTG}. Like CounTR, CounTX includes a transformer-based image encoder, $f_{\theta_{ViT}}$, and a transformer based feature interaction module, $f_{\varphi}$. While the feature interaction module in CounTR is used to mine the similarities between image patches and visual exemplars, the feature interaction module in CounTX is used to mine the similarities between image patches and class descriptions. In this way, CounTX is a natural extension from open-world object counting using image exemplars to open-world object counting using text descriptions.

For an image, $X$, with class description, $t$, an estimate, $\hat{y}$, of the number of objects described by $t$ in $X$ can be obtained from CounTX as:
\begin{equation}\label{mod_eq}
    \hat{Y} = f_{\psi}(f_{\varphi}(f_{\theta_{ViT}}(X), f_{\theta_{TT}}(t)))
\end{equation}
where  $\hat{Y}$ is the generated density map, with entries that sum to the object count (i.e., $\hat{y} = \sum_{p, q}(\hat{Y})_{p, q}$). Next, we describe each module in equation~\ref{mod_eq} in detail.

\paragraph{Image Encoder ($f_{\theta_{ViT}}$).}
For the image encoder, $f_{\theta_{ViT}}$, the CLIP vision transformer ViT-B-16 \cite{Radford2021LearningTV, ilharco_gabriel_2021_5143773} is used. This image backbone has been contrastively pretrained with the text encoder, $f_{\theta_{TT}}$, on image-text pairs in LAION-2B \cite{schuhmann2022laionb}. The ViT-B-16 model has a patch size of $16 \times 16$, 12 layers, and a final embedding dimension of 512. Only the patch tokens output by this image encoder are used, and the CLS token is discarded. 
\paragraph{Text Encoder ($f_{\theta_{TT}}$).}
For the text encoder, $f_{\theta_{TT}}$, the CLIP text transformer \cite{Radford2021LearningTV, ilharco_gabriel_2021_5143773} contrastively pretrained with the ViT-B-16 image encoder, $f_{\theta_{ViT}}$, on LAION-2B \cite{schuhmann2022laionb} is used. $f_{\theta_{TT}}$ has a context length of 77, 12 layers, and a final embedding dimension of 512. While $f_{\theta_{ViT}}$ transforms each input image into a spatial map of 512-dimensional feature vectors corresponding to the image patches, $f_{\theta_{TT}}$ transforms a class description into a single 512-dimensional feature vector. 
\paragraph{Feature Interaction Module ($f_{\varphi}$).}
To fuse the information captured by the image features $f_{\theta_{ViT}}(X)$ and the text feature $f_{\theta_{TT}}(t)$, two transformer decoder layers with embedding dimensions of 512 are used in $f_{\varphi}$. $f_{\varphi}$ uses the image features $f_{\theta_{ViT}}(X)$ to compute the query vectors and the text feature $f_{\theta_{TT}}(t)$ to compute the key and value vectors. The cross-attention mechanisms in $f_{\varphi}$ allow the model to leverage any similarities between the image patches and the class description preserved by $f_{\theta_{ViT}}(X)$ and $f_{\theta_{TT}}(t)$. 
\paragraph{Decoder ($f_{\psi}$).}
Before being passed to the decoder, $f_{\psi}$, the output of the feature interaction module, $f_{\varphi}$, is reshaped into a spatial feature map with 512 channels. Each channel is upsampled using bilinear interpolation to $24 \times 24$ pixels. The resized maps are then passed through a convolutional layer with 256 filters and upsampled to increase their height and width by a factor of two four times. This progressive four-block convolution and upsampling operation results in a feature map with the same height and width as the input image and 256 channels. These channels are combined into a single-channel density map using a $1\times1$ convolution. This density map is summed to estimate the object count. 
\subsection{Training Procedure}\label{train}
The image and text encoders were pre-trained on abundant image-text pairs using CLIP \cite{Radford2021LearningTV}. Therefore, prior to fine-tuning CounTX on the counting task, the image encoder and the text encoder map the input images and class descriptions to a joint text-image embedding space, aiding the feature interaction module in comparing data from the two modalities. Thus, the image and text encoders are first initialized with their pre-trained weights from CLIP. The text encoder is then frozen, while the image encoder is fine-tuned with the rest of the model on the counting task. As shown in section \ref{finetune_or_freeze}, this combination of fine-tuning and freezing the image and text encoders produces the best performance. The augmentation and scalable mosaicing schemes used in \cite{Liu2022CounTRTG} are employed during fine-tuning. The mean squared per-pixel error between the predicted density map, $\hat{Y}$, and the ground truth density map, $Y$, as shown in equation \ref{loss}, is averaged over all the images in each batch to compute the total loss for optimization.
\begin{equation}\label{loss}
    \mathcal{L}(\hat{Y}, Y) = \frac{1}{H \times W}\sum_{p, q}((\hat{Y})_{p, q} - (Y)_{p, q})^{2}
\end{equation}

\section{Experiments}

\subsection{Datasets \& Metrics}
\paragraph{Datasets.}
CounTX is evaluated on FSC-147~\cite{m_Ranjan-etal-CVPR21}, a class-agnostic object counting dataset containing 6135 images. The FSC-147 training set contains images from 89 classes, while the validation and test sets each contain images from 29 classes. The classes in the training, validation, and test sets are disjoint, making FSC-147 an open-world dataset. FSC-147 offers three visual exemplars for each image, but CounTX does not use them. 

\begin{figure*}[h!]
\centering
\includegraphics[width=\textwidth]{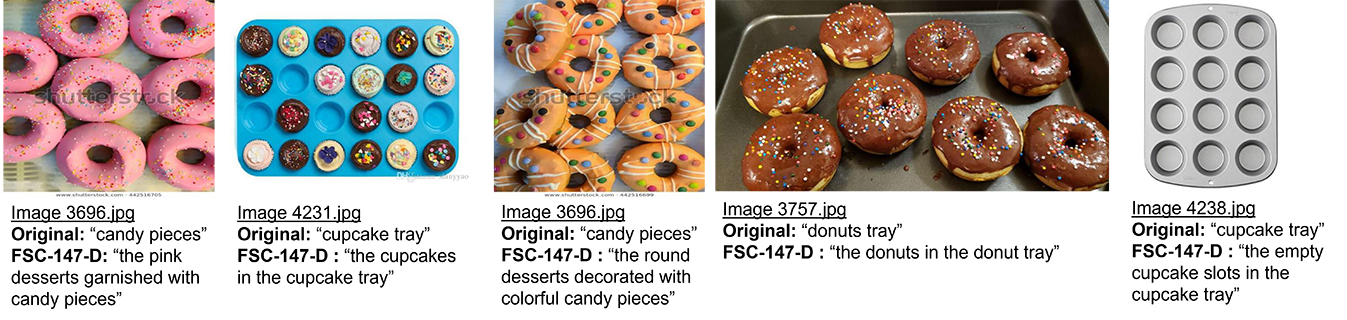}
\vspace{-7mm}
   \caption{Examples of changes made to FSC-147 to construct FSC-147-D, a dataset for the open-world text-specified object counting setting.}
\vspace{-3mm}
\label{fig:mod_fsc147}
\end{figure*}

While FSC-147 includes class names, these class names are not natural language responses to the question ``what object should be counted?'' Furthermore, some of the class names do not accurately describe the object being counted. We construct a set of descriptions  from FSC-147 suitable for our setting that transform the FSC-147 class names to responses to the question ``what object should be counted?'' and correct any mistakes that we find. We name this set of descriptions `FSC-147-D'. For instance, the original class name for image 3696.jpg in FSC-147 does not indicate that the pastries, not the candy pieces on top of the pastries, should be counted. The original class name for image 4231.jpg in FSC-147 is incorrect as the cupcakes are supposed to be counted, not the tray holding them. These class names are changed in FSC-147-D as shown in Figure \ref{fig:mod_fsc147}.

In addition to FSC-147, CounTX is evaluated qualitatively on a subset of CountBench \cite{paiss2023countclip}, a new text-image benchmark with a total of 540 images containing between two to ten instances of a particular object, where each image's caption reflects this number. CountBench is not a benchmark for text-specified object counting as its captions contain the number of objects to be counted. Thus, for qualitative evaluation, the CountBench captions were replaced with responses to the question ``what object should be counted?'' For example, the original CountBench caption for the leftmost image in Figure \ref{fig:count_bench_densities} is ``a set of six enamelled, gilt silver espresso spoons, tillander, helsinki 1955-56,'' which was replaced with ``the gilt silver espresso spoons'' for the text-specified object counting setting.

\paragraph{Metrics.} The Mean Absolute Error (MAE) and the Root Mean Squared Error (RMSE) are used to measure the performance of CounTX. The MAE and RMSE are given by: 
\begin{equation}\label{err_forms}
    MAE = \frac{1}{N}\sum_{i = 1}^{N}|\hat{y}_{i} - y_{i}|,\quad\text{ }RMSE = \sqrt{\frac{1}{N}\sum_{i = 1}^{N}(\hat{y}_{i} - y_{i})^2}
\end{equation}
where $N$ is the number of test images, $\hat{y}_{i}$ is the predicted count for image $X_{i}$, and $y_{i}$ is the ground truth count for image $X_{i}$.
\subsection{Implementation}
\paragraph{Training.} Each training image is cropped with a random square window with the same height as the original image and resized to $224 \times 224$ pixels. The images are then normalized before being passed through the model. To construct the ground truth density map for each image $X_{i}$, a Gaussian filter is applied to its corresponding binary map, $Y_{i}$ in Equation~\ref{count}, such that the filtered map still sums to the object count. For optimization, the loss defined in Equation~\ref{loss} averaged over each batch is minimized. Following \cite{Liu2022CounTRTG}, the density map values are scaled by 60, and errors contributed by pixels at certain positions are dropped with a 20 \% probability. Scaling by a factor of 60 prevents the model from generating a density map of zeros by increasing the penalty for this solution. CounTX is trained on images from the FSC-147 training set and text descriptions from FSC-147-D. We use the AdamW optimizer with $\beta_{1} = 0.9$ and $\beta_{2} = 0.95$, a batch size of 8, and a learning rate of $6.25 \times 10^{-6}$ that is warmed up for 10 epochs and then decayed with a half-cycle cosine schedule. The model at the epoch with the smallest mean absolute error on the validation set over 1000 epochs is selected.

\paragraph{Inference.} Following~\cite{Liu2022CounTRTG}, a square sliding window is scanned over the image with a stride of 128 pixels. As in training, each square sliding window of the image is resized to $224 \times 224$ pixels and normalized before being passed through the model. The density map for overlapping regions is computed using the averaging technique in~\cite{Liu2022CounTRTG}. 

\subsection{Comparison to State-of-the-art}
CounTX is evaluated on the FSC-147 dataset and compared against 0-shot exemplar-free methods that count instances of the dominant class, 3-shot exemplar-based methods, and text-based methods. As shown in Table \ref{sota}, CounTX trained on FSC-147-D achieves a new state-of-the-art performance across all measures on FSC-147 for text-specified object counting, significantly outperforming Patch-selection \cite{Xu2023ZeroshotOC}, both when evaluating with descriptions from FSC-147-D or with class names from the original FSC-147.

\begin{table}[t!]
\begin{center}
\scriptsize
\begin{NiceTabular}{|c|c|c|c|c|c|c|c|} 
\CodeBefore
    \rowcolors{10-11}{lightgray!30}{lightgray!30}[restart]
\Body
 \hline
 Method & Year & Published & How to Specify & \multicolumn{2}{c}{Validation} & \multicolumn{2}{c}{Test} \\
  &  &  & the Class & MAE & RMSE & MAE & RMSE \\

  \hline
  RepRPN-Counter \cite{10.1007/978-3-031-26316-3_5} & 2022 & \cmark & None & 31.69 & 100.31 & 28.32 & 128.76 \\
  RCC \cite{Hobley2022LearningTC} & 2022 & \xmark & None & 20.39 & 64.62 & 21.64 & 103.47\\
  CounTR (0-shot) \cite{Liu2022CounTRTG} & 2022 & \cmark & None & 17.40 & 70.33 & 14.12 & 108.01\\
  LOCA (0-shot) \cite{low_shot} & 2022 & \xmark & None & 17.43 & 54.96 & 16.22 & 103.96\\

  \hline
    Patch-selection \cite{Xu2023ZeroshotOC} & 2023 & \cmark & Text (class name) & 26.93 & 88.63 & 22.09 & 115.17\\
  \textbf{CounTX (FSC-147-D)} & \textbf{2023} & \textbf{-} & \textbf{Text (class name)} & 17.70 & \textbf{63.61} & \textbf{15.73} & 106.88\\
  \textbf{CounTX (FSC-147-D)} & \textbf{2023} & \textbf{-} & \textbf{Text (FSC-147-D)} & \textbf{17.10} & 65.61 & 15.88 & \textbf{106.29}\\
  \hline

  CounTR (3-shot) \cite{Liu2022CounTRTG} & 2022 & \cmark & 3 Visual Exemplars & 13.13 & 49.83 & 11.95 & 91.23\\
  LOCA (3-shot) \cite{low_shot} & 2022 & \xmark & 3 Visual Exemplars & 10.24 & 32.56 & 10.79 & 56.97\\
  
 \hline
\end{NiceTabular}
\caption{\label{sota}State-of-the-art performance on FSC-147 for exemplar-free, exemplar-based, and text-based methods. Note that CounTX is trained on FSC-147-D, but evaluated under two different settings: specifying classes with class names (from the original FSC-147) or with descriptions (from FSC-147-D). CounTR and LOCA are grayed out because they use visual exemplars, which provide more information than class descriptions.}
\end{center}
\vspace{-4mm}
\end{table}

\subsection{Ablation Study}

\paragraph{Freezing vs.\ Fine-tuning.}\label{finetune_or_freeze}
Different freezing and fine-tuning strategies are compared for both the image encoder and the text encoder. As shown in Table \ref{freeze_or_finetune}, freezing the text encoder and fine-tuning the image encoder on the counting task results in the best performance on FSC-147 across all measures.

\begin{table}[h!]
\begin{center}
\scriptsize
\begin{tabular}{|c|c|c|c|c|c|} 
 \hline
 Image Encoder & Text Encoder & \multicolumn{2}{c|}{Validation} & \multicolumn{2}{c|}{Test} \\
  Frozen &  Frozen & MAE & RMSE & MAE & RMSE \\
  \hline
  Yes & Yes & 37.92 & 103.57 & 37.37 & 130.36\\
  Yes & No & 33.34 & 100.58 & 37.61 & 131.2\\
  No & No & 17.73 & 68.19 & 16.39 & 107.65\\
  \textbf{No} & \textbf{Yes} & \textbf{17.10} & \textbf{65.61} & \textbf{15.88} & \textbf{106.29}\\
 \hline
\end{tabular}
\caption{\label{freeze_or_finetune} Performance of different freezing and fine-tuning strategies on FSC-147.}
\end{center}
\end{table}

\subsection{Qualitative Results}
\begin{figure*}[h!]
\begin{center}
\includegraphics[width=1\textwidth]{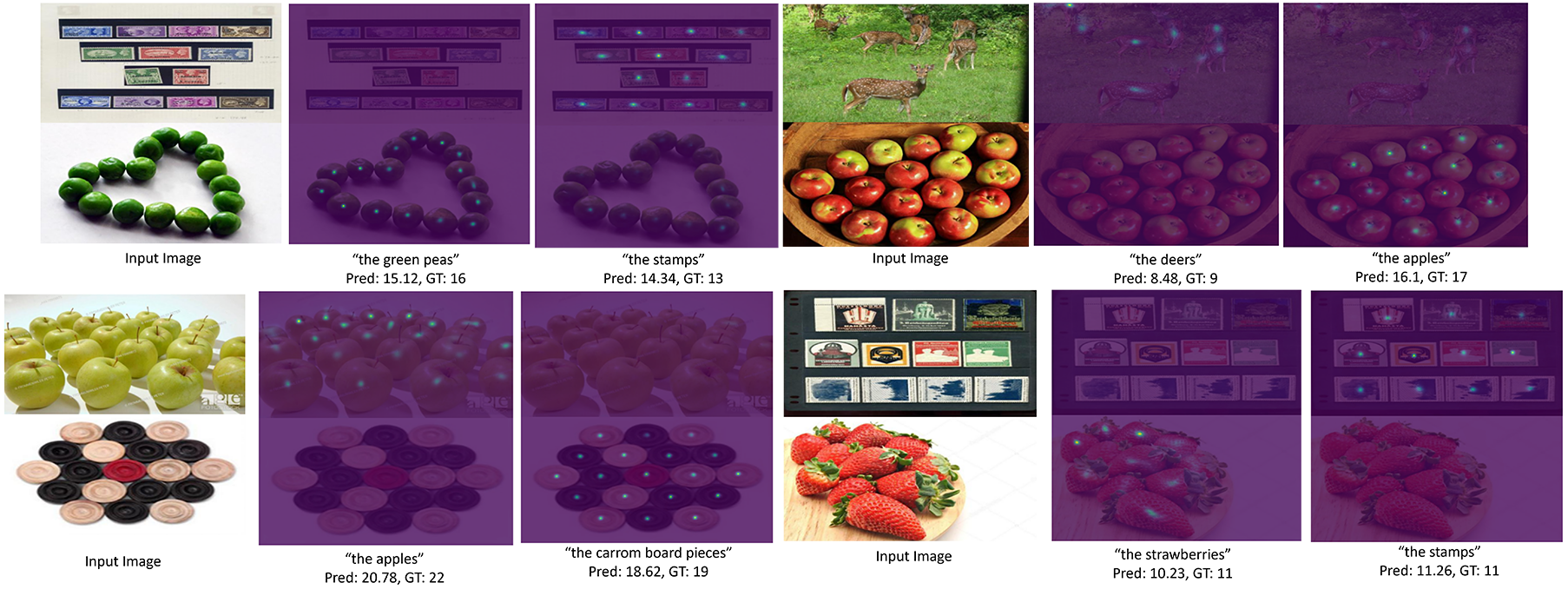}
\end{center}\vspace{-5mm}
   \caption{Evaluation on composite images. CounTX uses the class description to identify the object to count. This is clear by how the density map highlights only the regions specified by the class description in each example.}
\label{fig:mosaic}
\vspace{-3mm}
\end{figure*}

\paragraph{FSC-147 Test Set Image Mosaics.}To verify that CounTX uses the class description to count objects, pairs of images in the FSC-147 test set are stitched together, and CounTX is tasked to predict the counts of different classes in the same mosaicked image. In Figure~\ref{fig:mosaic}, a few examples of when the model clearly distinguished between classes using the class description are presented.

\paragraph{CountBench Density Maps.} To further investigate CounTX's generalization abilities to images with small numbers of class instances, responses to the question ``what object should be counted?'' were constructed for a subset of CountBench \cite{paiss2023countclip}, and CounTX was applied to this subset. A few of the density maps generated by CounTX are included in Figure \ref{fig:count_bench_densities} and more examples appear in the supplementary material.
\begin{figure*}[h!]
\centering
\includegraphics[width=1\textwidth]{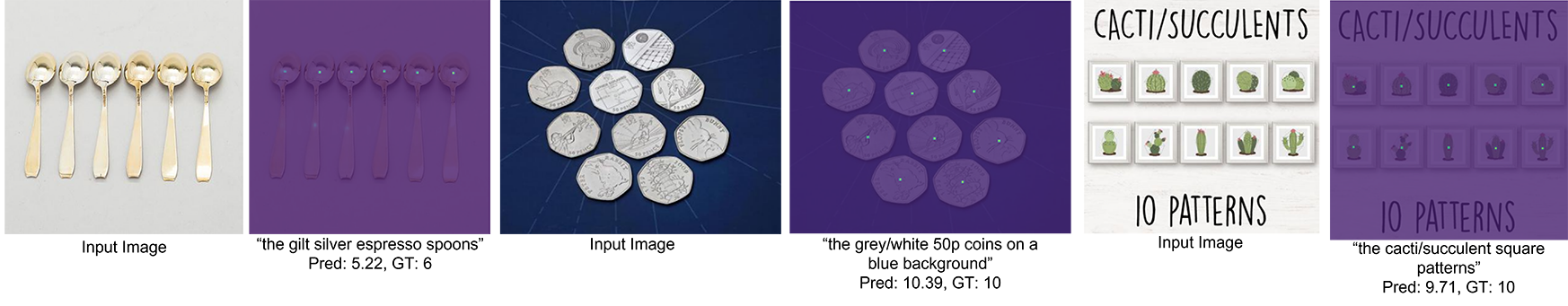}
\vspace{-5mm}
   \caption{Density maps produced by CounTX when applied to the CountBench subset.}
\label{fig:count_bench_densities}
\end{figure*}

\begin{figure*}[h!]
\begin{center}
\includegraphics[width=0.93\textwidth]{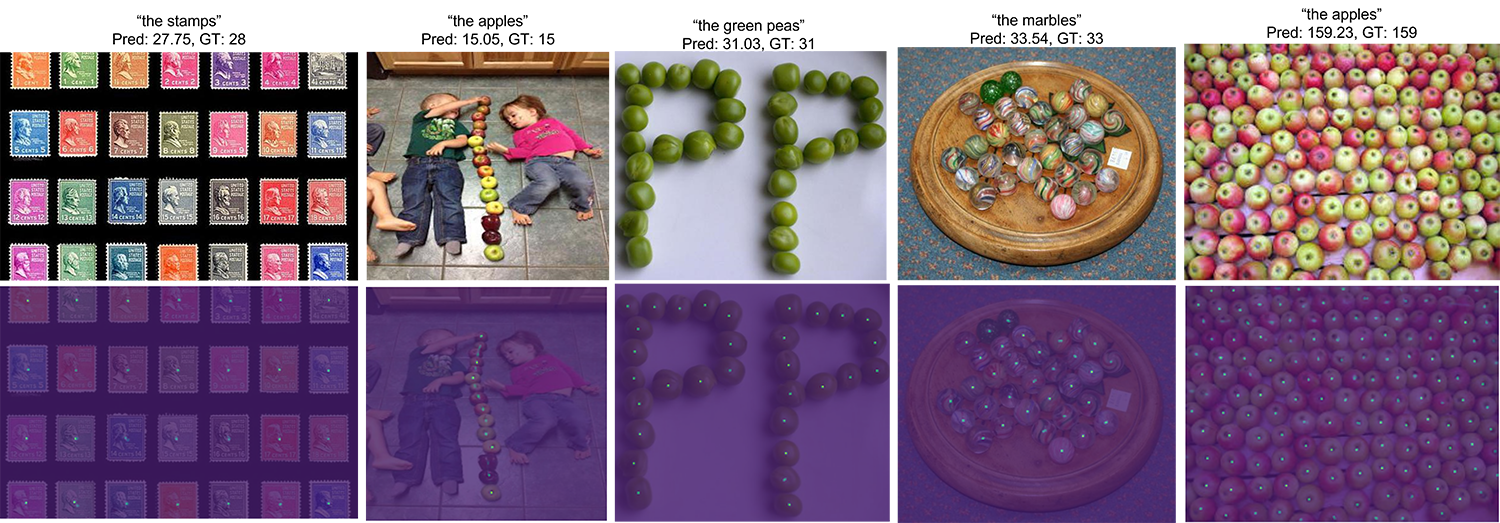}
\includegraphics[width=0.93\textwidth]{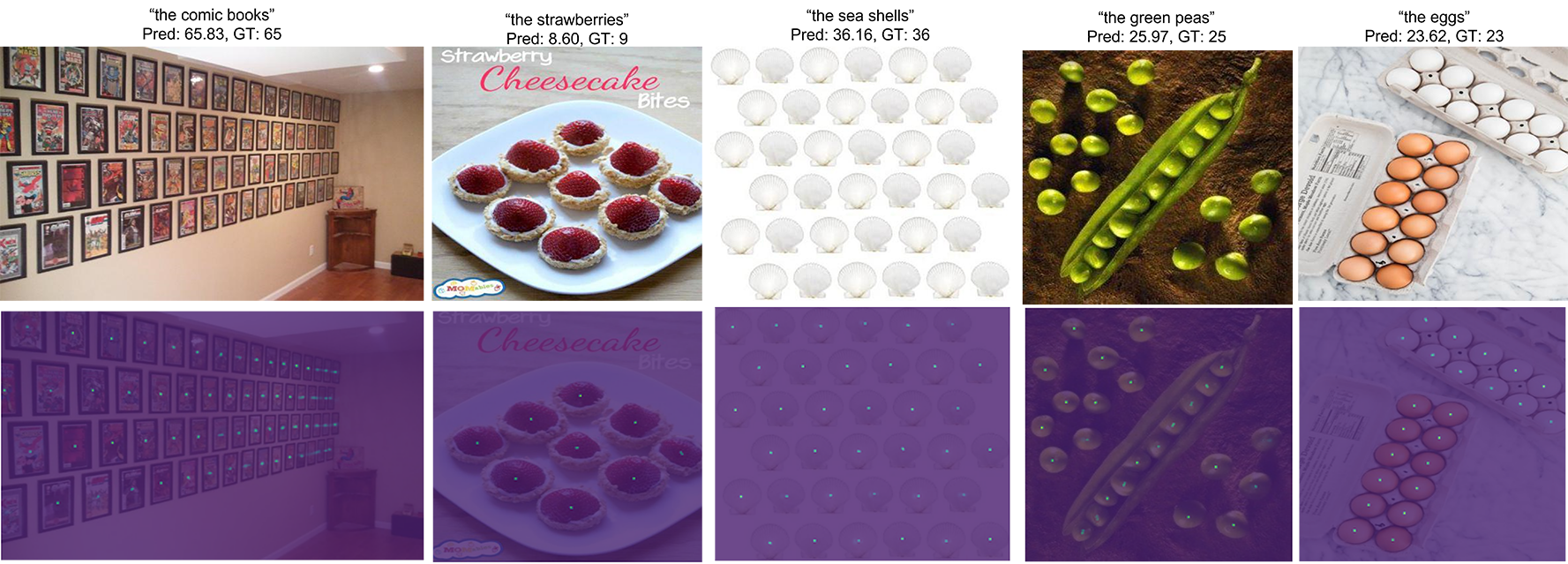}
\end{center}\vspace{-6mm}
   \caption{Density maps produced by CounTX when applied to the FSC-147 test set.}
\label{fig:test_set_densities}
\end{figure*}

\paragraph{FSC-147 Test Set Density Maps.} In Figure \ref{fig:test_set_densities}, examples of density maps produced by CounTX when applied to the FSC-147 test set are presented. Each density map is overlaid on top of its corresponding image. The class description, predicted count, and ground truth count are also provided.
\section{Conclusion \& Future Work}
This paper proposes CounTX, an open-world object counting model that accepts an image and an \emph{arbitrary} object class description, and directly uses these inputs to predict the object count. The paper also presents FSC-147-D, an augmented version of FSC-147 (a standard benchmark for class-agnostic counting) with class descriptions. Trained on FSC-147-D, CounTX demonstrates state-of-the-art performance across all measures on FSC-147 for methods that use text to specify the class. 

It has been mentioned multiple times in the literature \cite{lit, oquab2023dinov2, paiss2023countclip} that CLIP models, such as the image encoder used for CounTX, lack the spatial awareness needed for tasks such as counting. This is because, during their pre-training, features from CLIP models may not need to capture the rich structural information in images. Therefore, future work would include replacing the image encoder in CounTX with a more generally trained image backbone that also maps images to a joint text-image embedding space. Some possible visual backbones include a LiT DinoV2 image encoder \cite{lit, oquab2023dinov2}, or a LiT vision transformer trained with self-supervised patch reconstruction to improve its spatial awareness. The CounTX framework together with these spatially aware image encoders could be extended to models that answer other visual questions that require an understanding of spatial layout. These might include queries about the area, shape, and structure of objects in a scene.

\subsubsection*{Acknowledgement}
The authors would like to thank Chang Liu for his extensive support of the CounTR implementation, and Weidi Xie for insightful discussions. T.\ Han would like to thank Yuki M Asano and Lukas Knobel for helpful discussions. This research is funded by an AWS Studentship, the Reuben Foundation, the AIMS CDT program at the University of Oxford, EPSRC Programme Grant VisualAI EP/T028572/1, and a Royal Society Research Professorship RP \textbackslash R1 \textbackslash 191132.
\bibliography{egbib}
\newpage

\vspace{10mm}
\section*{Appendix}
\appendix
Section~\ref{sec:add_imp_details} describes 
additional implementation details about the CounTX training algorithm. Section \ref{sec:additional_datasets} presents and analyzes CounTX's performance on additional datasets. Section~\ref{sec:image_encoder_backbones} compares the quality of different frozen image encoder backbones for the counting task. Further information about FSC-147-D is provided in section~\ref{sec:fsc-147-d}. Section~\ref{sec:qual} illustrates additional density maps generated by CounTX to supplement the images already included in the paper. Finally, section~\ref{sec:failures} discusses known weaknesses of CounTX to be improved on in future work.
\section{Additional Training Implementation Details}
\label{sec:add_imp_details}

In this section, additional implementation details about the CounTX augmentation scheme and the construction of the ground truth density maps are discussed. During training, images are augmented with a probability of $\frac{2}{5}$. If augmentation is applied, either the augmentation pipeline presented in Table~\ref{aug} is used with a probability of $\frac{3}{8}$, or a scalable mosaicking scheme is employed with a probability of $\frac{5}{8}$. For the scalable mosaicking scheme, if an image contains greater than or equal to seventy objects to be counted, the same image is cropped four times to create the mosaicked image. Otherwise, four different training images are used. Cropping and combining the same image means the number of objects can be increased. Cropping and combining four different images teaches the model to distinguish between different semantic categories using the text descriptions. $\alpha$-channel blending is applied to soften the sharp borders between the four different crops in the mosaicked image. These augmentation techniques were adopted from CounTR \cite{Liu2022CounTRTG}, which can be referenced for further details. To construct the ground truth density maps, the provided annotations with ones at the centers of the objects to be counted and zeros elsewhere were filtered with a Gaussian kernel with x and y standard deviations of one and a radius of four.
\begin{table}[h!]
\scriptsize
\centering
\begin{tabular}{|l|l|} 
  \hline
  \textbf{Augmentation} & \textbf{Settings} \\
   \hline
   Gaussian Noise & mean: 0\\ & standard deviation: 0.1 \\
   \hline
   Color Jitter & brightness factor: 0.25 \\ & contrast factor: 0.15 \\ & saturation factor: 0.15 \\ & hue factor: 0.15\\
  \hline
  Gaussian Blur & kernel size: (7, 9) \\ & standard deviation: sampled uniformly from $[0.1, 2]$\\
  \hline
  Random Affine & rotation: sampled uniformly from $[-15^{\circ}, 15^{\circ}]$\\ & scale factor: sampled uniformly from $[0.8, 1.2]$\\ & translation factor (x, y): sampled uniformly from $[-0.2, 0.2] \times [-0.2, 0.2]$\\ & shear (x, y): sampled uniformly from $[-10^{\circ}, 10^{\circ}] \times [-10^{\circ}, 10^{\circ}]$\\
  \hline
  Horizontal Flip & horizontally flipped with probability $\frac{1}{2}$\\
  \hline
 \end{tabular}
 \vspace{1mm}
 \caption{\label{aug} CounTX augmentation pipeline. The augmentations are applied during training with a probability of $\frac{3}{20}$ in the top-to-bottom order of the rows in the table.}
 \end{table}
 
 \newpage
\section{Additional Experiments on Other Datasets}
\label{sec:additional_datasets}

\subsection{Val-COCO \& Test-COCO}
A straightforward approach to object counting is to enumerate all the class instances produced by pre-trained object detectors such as RetinaNet \cite{retina-net} and Faster-RCNN \cite{faster-rcnn} or by an instance segmentation model such as Mask-RCNN \cite{mask-rcnn}. Therefore, it is instructive to investigate how CounTX performs compared to these models. However, unlike CounTX, RetinaNet, Faster-RCNN, and Mask-RCNN are closed-set methods and, thus, are limited to counting instances of classes they were trained on. On the other hand, CounTX is an open-set model and, as a result, can count instances of arbitrary classes.\\

The FSC-147 \cite{m_Ranjan-etal-CVPR21} dataset provides Val-COCO and Test-COCO, image subsets of the COCO \cite{Lin2014MicrosoftCC} dataset. RetinaNet, Faster-RCNN, and Mask-RCNN have been trained to categorize objects into the classes present in these subsets. As a result, CounTX was evaluated against these methods using Val-COCO and Test-COCO. As shown in table \ref{coco}, CounTX performs better than all three closed-set methods and the class-agnostic counting model FamNet. However, unlike FamNet, CounTX does not require any visual exemplars for inference.

\begin{table}[t!]
\begin{center}
\scriptsize
\begin{NiceTabular}{|c|c|c|c|c|c|c|} 
\CodeBefore
    \rowcolors{7-10}{lightgray!30}{lightgray!30}[restart]
\Body

\hline
Method & Method Type & How to Specify & \multicolumn{2}{c}{Val-COCO} & \multicolumn{2}{c}{Test-COCO} \\
&  & the Class & MAE & RMSE & MAE & RMSE \\

\hline
RetinaNet \cite{retina-net} & Closed-set & Text (class name) & 63.57 & 174.36 & 52.67 & 85.86\\
Faster-RCNN \cite{faster-rcnn} & Closed-set & Text (class name) & 52.79 & 172.46 & 36.20 & 79.59 \\
Mask-RCNN \cite{mask-rcnn} & Closed-set & Text (class name) & 52.51 & 172.21 & 35.56 & 80.00\\
\hline
    
\textbf{CounTX (FSC-147-D)} & Open-set & \textbf{Text (FSC-147-D)} & \textbf{29.39} & \textbf{101.56} & \textbf{12.15} & \textbf{25.49}\\
\hline

FamNet \cite{m_Ranjan-etal-CVPR21} & Open-set & Visual Exemplars & 39.82 & 108.13 & 22.76 & 45.92\\
BMNet+ \cite{Shi2022RepresentCA} & Open-set & Visual Exemplars & 26.55 & 93.63 & 12.38 & 24.76\\
CounTR \cite{Liu2022CounTRTG} & Open-set & Visual Exemplars & 24.66 & 83.84 & 10.89 & 31.11\\
LOCA \cite{low_shot} & Open-set & Visual Exemplars &  16.86 & 53.22 & 10.73 & 31.31\\
\hline
  
\end{NiceTabular}
\caption{\label{coco} Performance of closed-set and open-set models on the Val-COCO and Test-COCO subsets of COCO \cite{Lin2014MicrosoftCC} and FSC-147 \cite{m_Ranjan-etal-CVPR21}. Methods in the bottom four rows are grayed out because they use visual exemplars, which provide more information than class descriptions.}
\end{center}
\vspace{-4mm}
\end{table}

\subsection{CARPK}
CounTX is evaluated quantitatively and qualitatively on the CARPK \cite{retina-net} dataset to demonstrate its ability to generalize to datasets other than FSC-147 \cite{m_Ranjan-etal-CVPR21}. The CARPK dataset for counting cars contains overhead images of parking lots captured by drone cameras. The CARPK training set includes 989 images, and the CARPK test set includes 459 images. CARPK contains photos of 90,000 cars altogether.\\

CounTX was trained and evaluated in multiple settings for CARPK \cite{retina-net} as shown in Table \ref{carpk}. For the fifth and sixth rows in Table \ref{carpk}, CounTX was trained on FSC-147 \cite{m_Ranjan-etal-CVPR21} and evaluated on the CARPK test set. For the seventh and eight rows in Table \ref{carpk}, CounTX was jointly trained on data from FSC-147 and CARPK and evaluated on the CARPK test set. Specifically, during joint training, each batch was constructed from data from either FSC-147 or CARPK. The batches were composed and shuffled randomly, and augmentation was only applied to data from FSC-147. Table \ref{carpk} illustrates CounTX's performance using different potential responses to the query ``what object should be counted'' for CARPK, as indicated by the third column.\\

\begin{figure*}[h!]
\begin{center}
\includegraphics[width=0.8\textwidth]{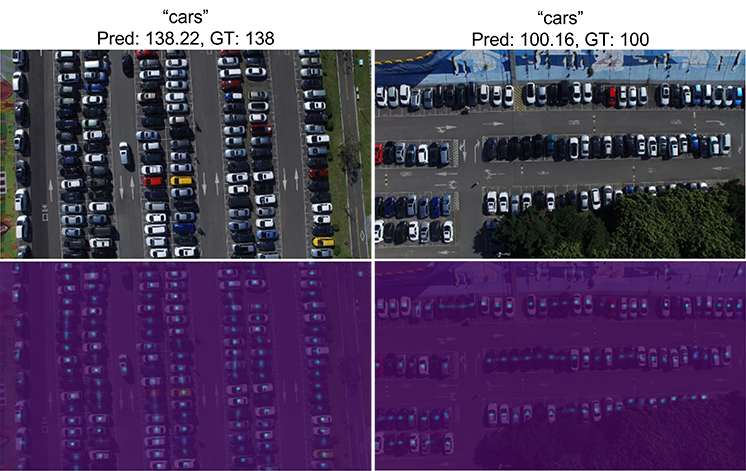}
\end{center}
   \caption{Density maps produced by CounTX, trained on FSC-147 \cite{m_Ranjan-etal-CVPR21}, when applied to the CARPK \cite{retina-net} test set with no fine-tuning.}
\label{fig:carpk}
\end{figure*}

As shown in Table \ref{carpk}, CounTX performs competitively compared to closed-set counting methods on CARPK \cite{retina-net}. Methods with asterisks in Table \ref{carpk} were trained specifically for car detection, while the other closed-set methods can count instances of other classes. With and without being trained on data in CARPK, CounTX performs better than the few-shot class-agnostic counting method FamNet \cite{m_Ranjan-etal-CVPR21} trained on data from CARPK. It is interesting to consider whether training FamNet on CARPK damages its performance on FSC-147 \cite{m_Ranjan-etal-CVPR21} significantly. Similarly, the fine-tuning of models such as CounTR \cite{Liu2022CounTRTG} and SAFECount \cite{You_2023_WACV} could cause them to lose their generality. The joint training procedure for CounTX avoids this issue by ensuring that the model performs well on both FSC-147 and CARPK during the final optimization process. The performance of CounTR and SAFECount pre-trained on FSC-147 and then fine-tuned on CARPK is shown in the bottom two rows of Table \ref{carpk}. Figure \ref{fig:carpk} illustrates the effectiveness and generality of CounTX trained only on data from FSC-147 and evaluated on the CARPK test set.
\begin{table}[t!]
\begin{center}
\scriptsize
\begin{NiceTabular}{|c|c|c|c|c|} 
\CodeBefore
    \rowcolors{11-14}{lightgray!30}{lightgray!30}[restart]
\Body

\hline
Method & Method Type & How to Specify & \multicolumn{2}{c}{CARPK} \\
&  & the Class & MAE & RMSE \\

\hline
Faster-RCNN \cite{faster-rcnn} & Closed-set & Text (class ``car'') & 39.88 & 47.67 \\
One-look Regression* \cite{Mundhenk2016ALC} & Closed-set & Text (class ``car'') & 21.88 & 36.73 \\
RetinaNet \cite{retina-net} & Closed-set & Text (class ``car'') & 16.62 & 22.30\\
HLCNN* \cite{Kili2021AnAC} & Closed-set & Text (class ``car'') & 2.12 & 3.02\\
\hline

CounTX (FSC-147) & Open-set & Text (description ``cars'') & 11.72 & 14.86\\
CounTX (FSC-147) & Open-set & Text (description ``car'') & 11.64 & 14.85\\
CounTX (FSC-147 \& CARPK) & Open-set & Text (description ``cars'') & 8.89 & 11.42\\
\textbf{CounTX (FSC-147 \& CARPK)} & Open-set & \textbf{Text (description ``the cars'')} & \textbf{8.13} & \textbf{10.87}\\
\hline

FamNet (CARPK) \cite{m_Ranjan-etal-CVPR21} & Open-set & Visual Exemplars & 18.19 & 33.66\\
CounTR (FSC-147 \& CARPK) \cite{Liu2022CounTRTG} & Open-set & Visual Exemplars & 5.75 & 7.45\\
SAFECount (FSC-147 \& CARPK) \cite{You_2023_WACV} & Open-set & Visual Exemplars & 5.33 & 7.04\\
\hline
  
\end{NiceTabular}
\caption{\label{carpk} Performance of closed-set and open-set models on the CARPK \cite{retina-net} dataset. Methods in the bottom three rows are grayed out because they use visual exemplars, which provide more information than class descriptions. Methods with asterisks were trained specifically for car counting, while other methods can count objects of other classes as well.}
\end{center}
\vspace{-4mm}
\end{table}

\newpage
\section{Additional Ablation Study: Image Encoder Backbones}
\label{sec:image_encoder_backbones}

To measure the quality of different image features for object counting, three different CLIP image encoder backbones were frozen and used to train CounTX with visual exemplars instead of text. The CounTX {\em text} encoder was replaced with the exemplar encoder (4 convolutional layers followed by global average pooling) from CounTR~\cite{Liu2022CounTRTG}. The training and inference procedures from CounTR were also adopted. As shown in Table~\ref{backbone}, compared to the other two CLIP models, the image encoder used in the main paper for CounTX, ViT-B-16, performs competitively on FSC-147~\cite{m_Ranjan-etal-CVPR21}. 

The image encoder from CounTR \cite{Liu2022CounTRTG} (pre-trained on ImageNet and then with self-supervised patch reconstruction on FSC-147 \cite{m_Ranjan-etal-CVPR21}) was also frozen and evaluated. It has been mentioned multiple times in the literature \cite{lit, oquab2023dinov2} that CLIP image encoders may not provide as rich spatial features out-of-the-box as other more generally trained image backbones. This point is consistent with the results in Table~\ref{backbone}, as the frozen CounTR image encoder, pre-trained with self-supervision, performs generally better than all three CLIP image encoders on FSC-147. However, the CounTR image encoder backbone does not have an available joint text-image embedding space as the CLIP image encoder backbones do. 

\begin{table}[h!]
\scriptsize
\begin{tabular}{|c|c|c|c|c|c|c|c|} 
  \hline
  Image Encoder & Pre-training & Embedding & Spatial Feature & \multicolumn{2}{|c|}{Validation} & \multicolumn{2}{|c|}{Test} \\
   Backbone &  Method &  Dimension &  Map Shape & MAE & RMSE & MAE & RMSE \\
   \hline
   CounTR \cite{Liu2022CounTRTG} & ImageNet and SSL & 512 & $24 \times 24$ & 15.53 & 53.01 & 14.93 & 94.38 \\
   \hline
   RN50x16 \cite{ilharco_gabriel_2021_5143773} & YFCC100M Subset \cite{Radford2021LearningTV} & 768 & $12 \times 12$ & 32.84 & 98.37 & 26.96 & 100.31\\
   ViT-L-14-336 \cite{ilharco_gabriel_2021_5143773} & YFCC100M Subset \cite{Radford2021LearningTV} & 768 & $24 \times 24$ & 27.35 & 81.73 & \textbf{22.72} & 96.34\\
   \textbf{ViT-B-16 \cite{ilharco_gabriel_2021_5143773}} & \textbf{LAION-2B \cite{schuhmann2022laionb}} & \textbf{512} & $\mathbf{14 \times 14}$ & \textbf{26.37} & \textbf{71.28} & 24.96 & \textbf{91.64}\\
  \hline
 \end{tabular}
 \caption{\label{backbone} Performance of different frozen image encoder backbones on the FSC-147 \cite{m_Ranjan-etal-CVPR21} 3-shot visual exemplar counting task. The last three rows contain data from CLIP image encoder backbones, while the first row contains data from the CounTR image encoder backbone. SSL stands for self-supervised learning.}
 \end{table}

\newpage
\section{Details of the FSC-147-D Dataset}
\label{sec:fsc-147-d}
\begin{figure*}[h!]
\begin{center}
\includegraphics[width=0.4\textwidth]{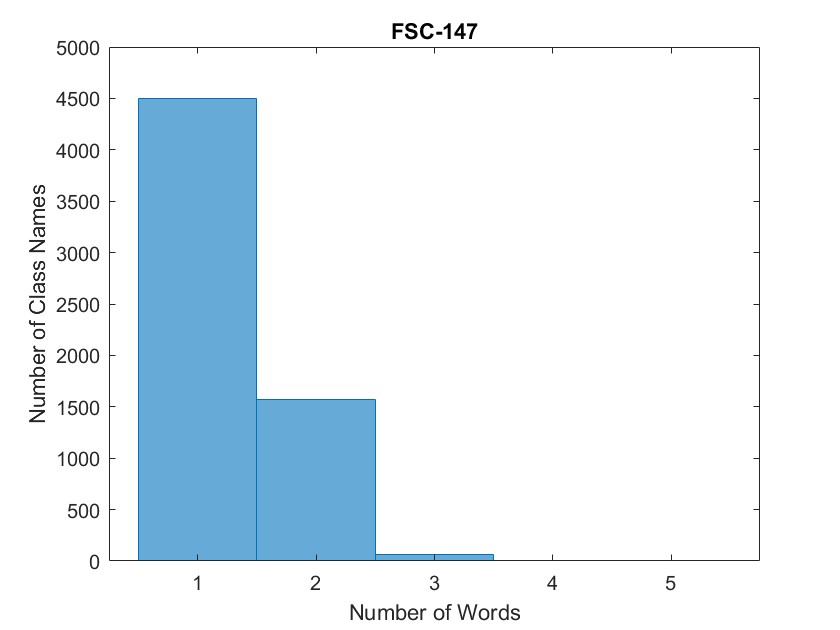}
\includegraphics[width=0.4\textwidth]{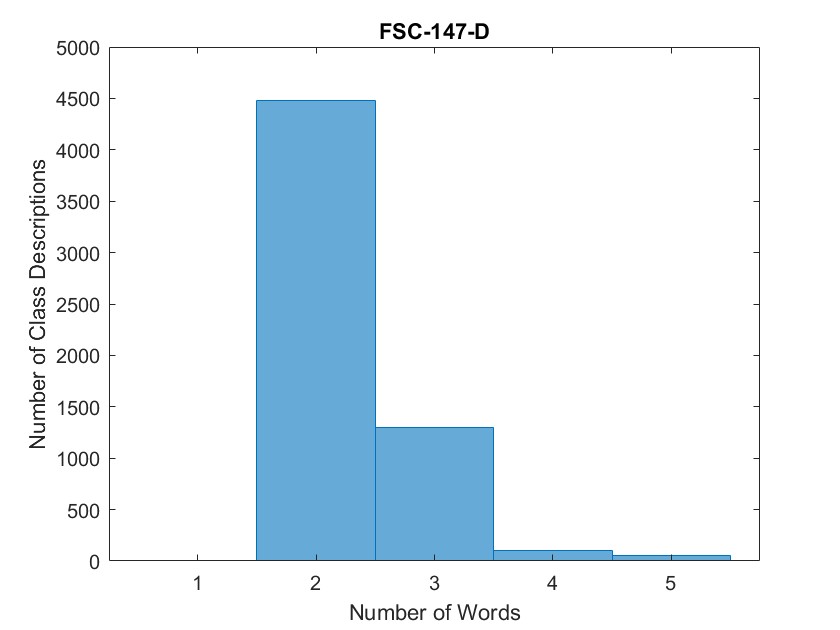}
\end{center}
   \caption{Histograms of the number of words in the class names for FSC-147 (left) and the number of words in the responses to the question ``what object should be counted?'' for FSC-147-D (right). The histograms show that the class descriptions in FSC-147-D are more prolific than the class names in FSC-147.}
\label{fig:fsc147_d_word_dist}
\end{figure*}

The file \texttt{FSC-147-D.json} contains the FSC-147-D dataset with class descriptions for the images in FSC-147 \cite{m_Ranjan-etal-CVPR21}. \texttt{FSC-147-D.json} is available at \\ \href{https://www.robots.ox.ac.uk/~vgg/research/countx/}{https://www.robots.ox.ac.uk/\textasciitilde vgg/research/countx/}.
While FSC-147 provides class names, FSC-147-D contains responses to the query ``what object should be counted?'' 92.4 \% of the class descriptions in FSC-147-D (5668 class descriptions) are the class names in FSC-147 with ``the'' prepended to them. The remaining 7.6 \% of the class names in FSC-147 (467 class names) required more complex rephrasing to convert them to class descriptions in FSC-147-D. Figure~\ref{fig:fsc147_d_word_dist} illustrates the distribution of the number of words for the class names in FSC-147 and the distribution of the number of words for the class descriptions in FSC-147-D.

\newpage
\section{Additional Counting Image Examples}
\label{sec:qual}
\subsection{FSC-147}
\label{sec:qual_fsc-147}
\begin{figure*}[h!]
\begin{center}
\includegraphics[width=0.9\textwidth]{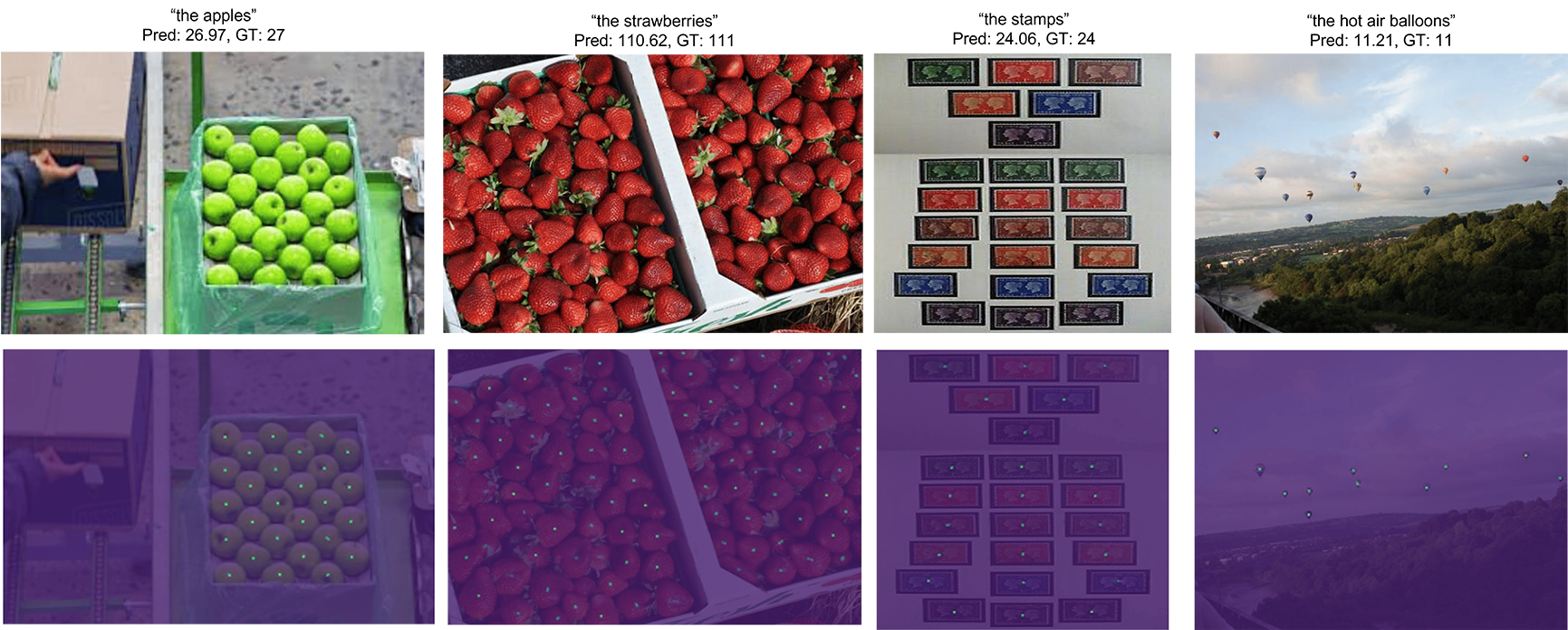}
\includegraphics[width=0.9\textwidth]{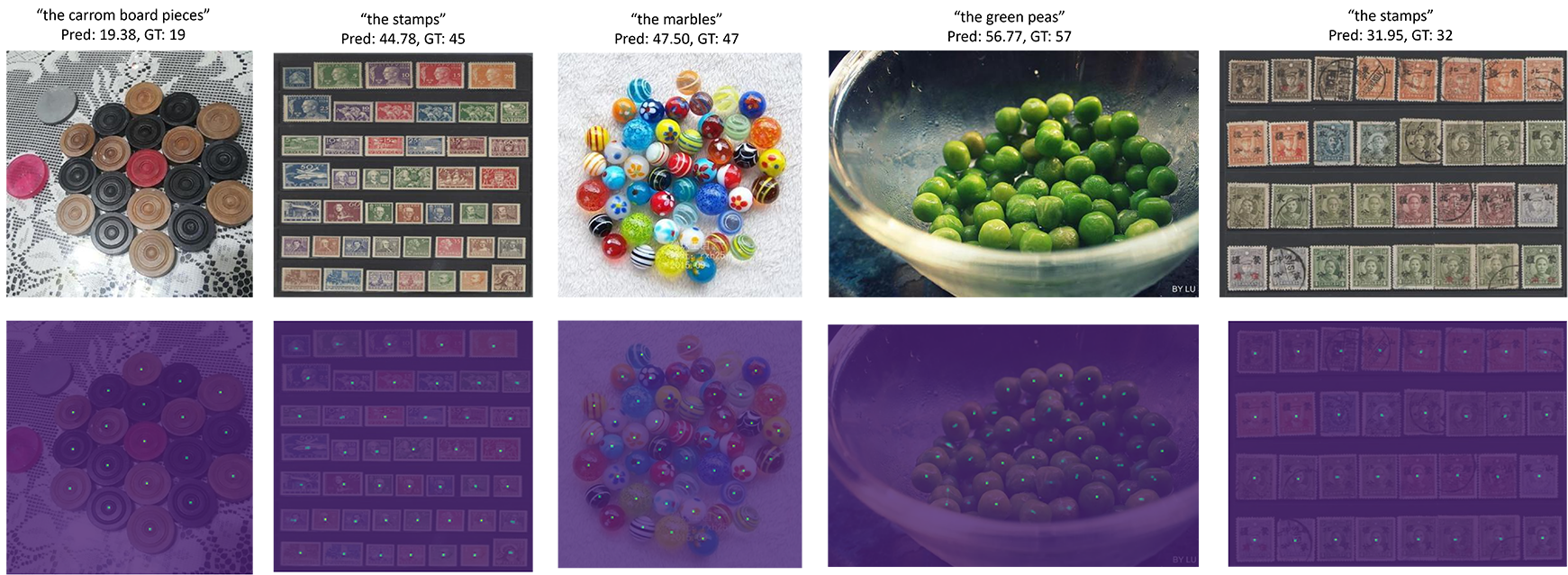}
\includegraphics[width=0.9\textwidth]{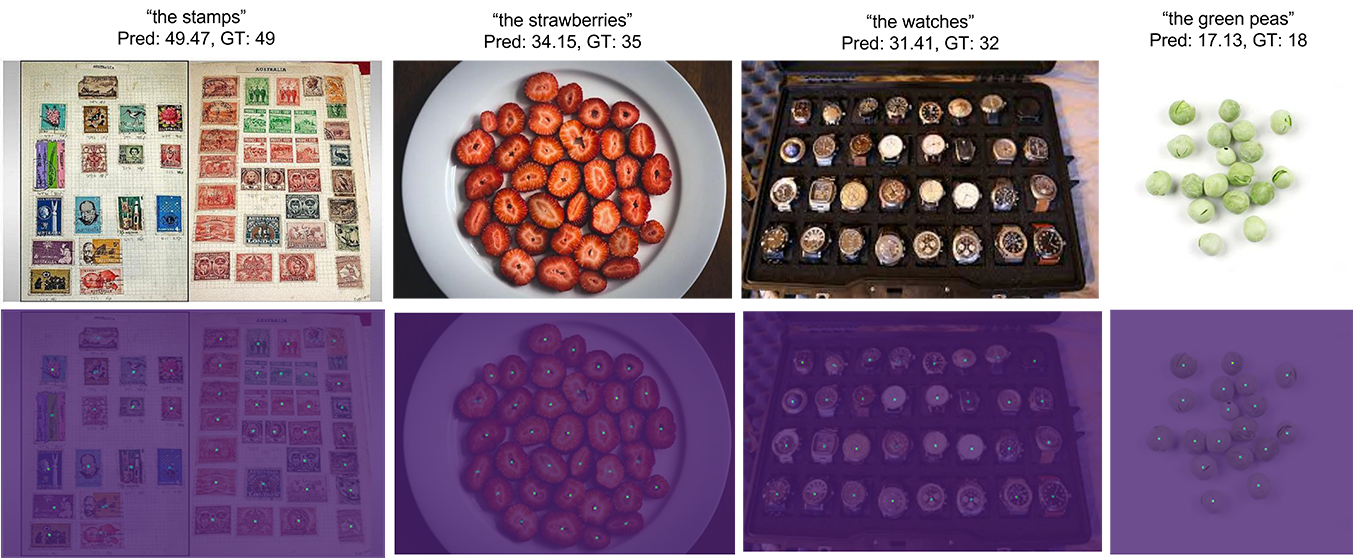}
\end{center}
   \caption{Density maps produced by CounTX when applied to the FSC-147 \cite{m_Ranjan-etal-CVPR21} test set. CounTX is able to count the hot air balloons in the rightmost image in the top row despite how small they are. CounTX also correctly counts only the carrom board pieces and excludes the extraneous circular objects in the leftmost image in the third row. Despite the variance in the color and shape of the stamps, CounTX counts them.}
\label{fig:fsc147_sup_qual_examples_1}
\end{figure*}

This section presents and comments on additional density maps produced by CounTX when applied to the FSC-147 \cite{m_Ranjan-etal-CVPR21} test set. Results are shown in Figure~\ref{fig:fsc147_sup_qual_examples_1}. More examples can be viewed at \href{https://www.robots.ox.ac.uk/~vgg/research/countx/}{https://www.robots.ox.ac.uk/\textasciitilde vgg/research/countx/}.

\newpage
\subsection{CountBench}
\label{sec:qual_countbench}
\begin{figure*}[h!]
\begin{center}
\includegraphics[width=0.8\textwidth]{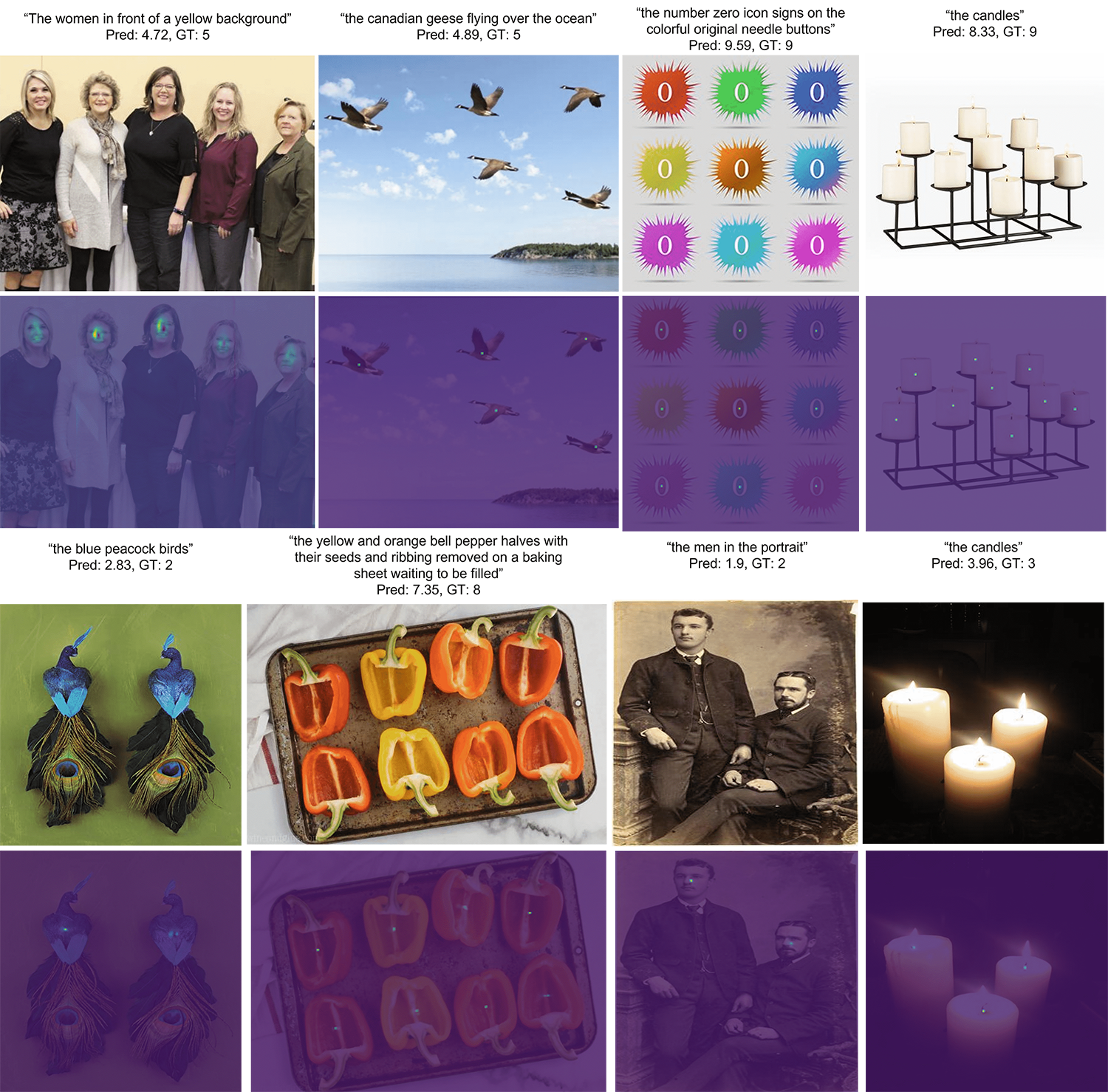}
\end{center}
   \caption{Density maps produced by CounTX when applied to the CountBench \cite{paiss2023countclip} subset. Even though CounTX was never trained to count people, it can count the women in the leftmost image in the top row and the men in the third image in the third row. CounTX estimates that there are almost exactly 2 men in the third image in the third row, even though no image in FSC-147 \cite{m_Ranjan-etal-CVPR21}, the dataset CounTX was trained on, has under 7 objects. The class descriptions in FSC-147-D are very simple compared to the long and detailed description of the bell peppers in the third row. Despite this, CounTX provides a reasonable estimate for the count of the bell pepper halves given this long description.}
\label{fig:countbench_sup_qual_examples_1}
\end{figure*}

Text descriptions were created for a subset of CountBench \cite{paiss2023countclip}. These descriptions are more detailed and longer in general than the descriptions in FSC-147-D. CountBench also only contains images with at most 10 objects. On the other hand, images in FSC-147 \cite{m_Ranjan-etal-CVPR21} contain at minimum 7 objects and at most 3731 objects with an average of 56 objects per image. Therefore, it is interesting to investigate how CounTX performs on the CountBench subset given that CounTX has never been trained on images with under 7 objects. In this section, qualitative examples are provided and commented on from such an investigation. Results are shown in Figure~\ref{fig:countbench_sup_qual_examples_1}. More examples can be viewed at \\
\href{https://www.robots.ox.ac.uk/~vgg/research/countx/}{https://www.robots.ox.ac.uk/\textasciitilde vgg/research/countx/}.

\newpage
\section{Limitations}
\label{sec:failures}
\begin{figure*}[h!]
\begin{center}
\includegraphics[width=1\textwidth]{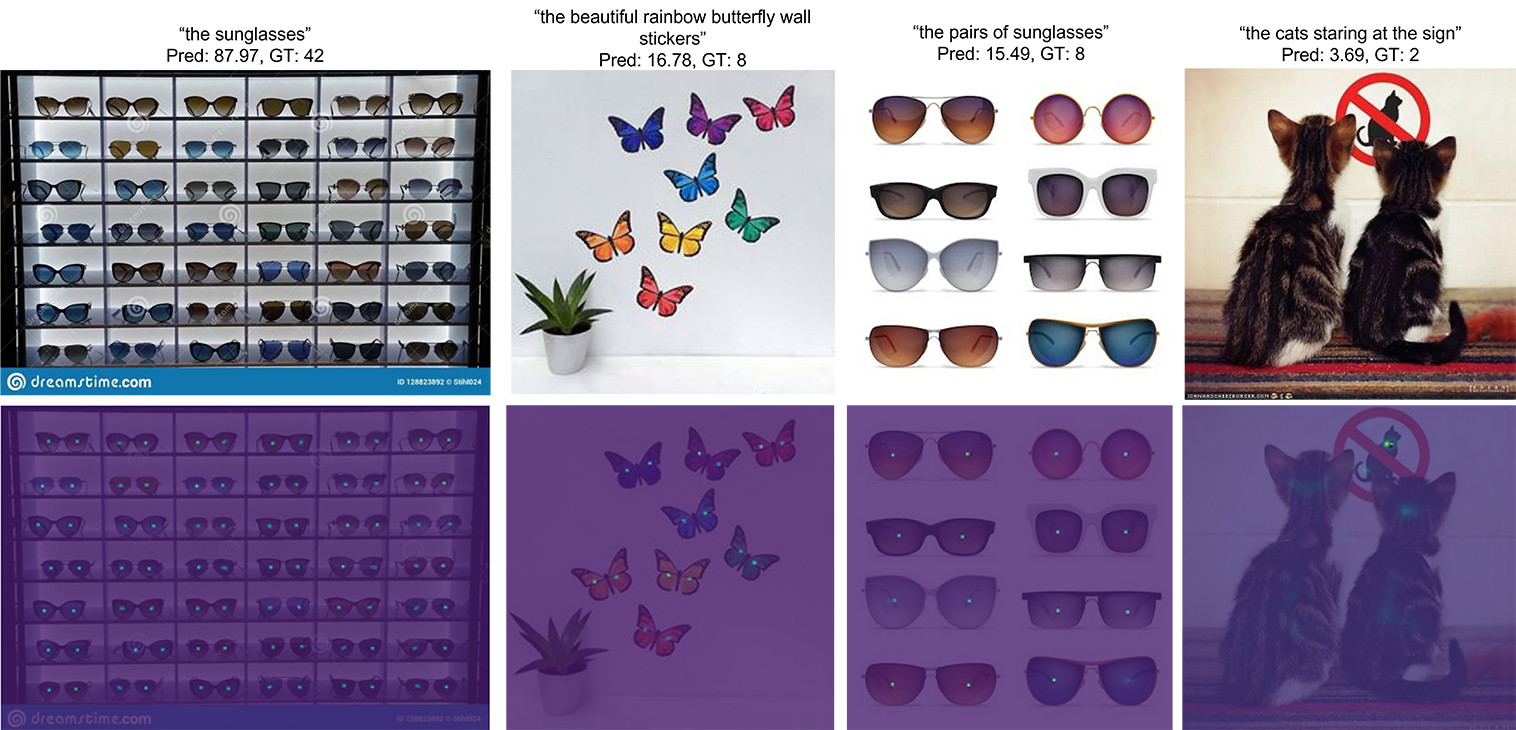}
\end{center}
   \caption{CounTX struggles to count self-similar objects. Instead of placing a dot on each pair of sunglasses in the density maps for the first image (from FSC-147 \cite{m_Ranjan-etal-CVPR21}) and the third image (from CountBench \cite{paiss2023countclip}), CounTX places a dot on each lens. This is why the estimated counts for these images are almost double the ground truth counts. Following this pattern, CounTX places a dot on each butterfly wing in the density map for the second image from CountBench above. This results in an estimated count that is almost twice the ground truth count. CounTX also struggles with inter-object relationships. This weakness surfaces when evaluating CounTX qualitatively on the CountBench subset as the text descriptions for images in the CountBench subset are more nuanced than the descriptions in FSC-147-D. In the rightmost image above from CountBench, CounTX incorrectly attempts to count all the cats in the image, real and illustrated, instead of just the real cats staring at the sign with an illustrated cat.}
\label{fig:failures}
\end{figure*}

In this section, two weaknesses of CounTX will be discussed. CounTX struggles when an object is self-similar. Instead of counting each self-similar object as a whole, CounTX might double count by placing a dot on each similar component in the density map. For example, a typical pair of sunglasses is self-similar because it is composed of two similar lenses. As shown in Figure \ref{fig:failures}, instead of counting each pair of lenses as a whole object, CounTX might count each lens in the pair as an individual object. If visual exemplars were available, the final count could be calibrated by dividing the estimated count by the average sum of the density map at each exemplar region. However, this is not currently possible with only text descriptions. Secondly, CounTX struggles to understand inter-object relationships. These weaknesses are illustrated and discussed in Figure~\ref{fig:failures}.

\newpage
\end{document}